\pdfoutput=1
\documentclass[final]{IEEEtran} 

\usepackage{amsmath,amsfonts,amssymb,bm,graphicx}

\begin{document}

\title{Likelihood-Based Semi-Supervised Model Selection with Applications to Speech Processing}

\author{Christopher M.~White,~\IEEEmembership{Member,~IEEE},
        Sanjeev P.~Khudanpur,~\IEEEmembership{Senior~Member,~IEEE}, \\ and
        Patrick J.~Wolfe,~\IEEEmembership{Senior~Member,~IEEE}

\thanks{C.~M.~White is with the Human Language Technology Center of Excellence (HLT-COE), Johns Hopkins University, and the Statistics and Information Sciences Laboratory, Harvard University, Oxford Street, Cambridge, MA 02138 (e-mail: cmwhite@seas.harvard.edu); S.~P.~Khudanpur is with the Center for Language and Speech Processing, Johns Hopkins University, Charles St., Baltimore, MD 21218 (email: khudanpur@jhu.edu); and P.~J.~Wolfe is with the Statistics and Information Sciences Laboratory, Harvard University (e-mail: wolfe@stat.harvard.edu).  This work was completed while C.~M.~White was an HLT-COE Graduate Fellow.}
}

\maketitle

\markboth
{Submitted Manuscript}
{Submitted Manuscript}

\begin{abstract}%
In conventional supervised pattern recognition tasks, model selection is typically accomplished by minimizing the classification error rate on a set of so-called development data, subject to ground-truth labeling by human experts or some other means.  In the context of speech processing systems and other large-scale practical applications, however, such labeled development data are typically costly and difficult to obtain.  This article proposes an alternative semi-supervised framework for likelihood-based model selection that leverages unlabeled data by using trained classifiers representing each model to automatically generate putative labels.  The errors that result from this automatic labeling are shown to be amenable to results from robust statistics, which in turn provide for minimax-optimal censored likelihood ratio tests that recover the nonparametric sign test as a limiting case.  This approach is then validated experimentally using a state-of-the-art automatic speech recognition system to select between candidate word pronunciations using unlabeled speech data that only potentially contain instances of the words under test.  Results provide supporting evidence for the utility of this approach, and suggest that it may also find use in other applications of machine learning.
\end{abstract}%


\ifCLASSOPTIONpeerreview
\begin{center}%
\bfseries EDICS Category: SPE-GASR
\end{center}
\fi
\IEEEpeerreviewmaketitle

\section{Introduction}
\label{sec:Introduction}

\IEEEPARstart{T}{his} article develops a simple and powerful
likelihood-ratio framework that enables the use of \emph{unlabeled}
development data for model selection and system optimization in the
context of large-scale speech processing.  Within the speech
engineering community, \emph{acoustic} likelihoods have long played a
prominent role both as a training criterion and an objective function to
aid in system development. Log-likelihood ratios have in turn featured
ever more prominently in areas such as speech, speaker, and language
recognition; for instance, it is now common practice that ``target''
model likelihoods are compared to those of a universal ``background''
model as part of many large-scale speech processing
systems~\cite{reynolds}.

\subsection{Model Selection Using Likelihood Ratios}

Comparing data likelihoods between competing models can serve as an effective means of model selection for classification
and regression tasks.   However, when
considering conditional likelihoods of the observed data given
\emph{labels} such as orthographic transcriptions of speech waveforms, previous work has assumed that orthographic labels have
been correctly assigned by human experts, and hence are known exactly.  However,
such ``labeled data'' do not come for free; their acquisition requires
the time and expertise of a trained linguist, hence limiting
scalability to the large sample sizes necessary to succeed
in practical speech engineering tasks.

This article thus posits a framework in which likelihoods
evaluated using labels that are \emph{automatically assigned} by two competing
systems can serve as proxies for likelihoods based on ground-truth
labeling.  This yields not only a methodologically sound algorithmic
framework through which to incorporate unlabeled data into the
likelihood-based model selection process, but also practical
engineering strategies for selecting between competing models in order
to optimize large-scale systems.  Experiments to select between
candidate word pronunciations in the context of state-of-the-art
speech processing systems, using well-known corpora and standard
metrics, serve to demonstrate the benefit of unlabeled development
data in the context of large-scale speech processing.

To construct this framework, insights from robust statistics are used
to formulate the resultant semi-supervised model selection problem in a manner that permits principled analysis, and from which efficient and effective algorithms can be derived.  By considering the automatic labeling procedure as a mixture of correct and incorrect assignments, the influence of incorrect
labeling can be limited through what is known as a \emph{censored}
likelihood ratio evaluation.

The well-known nonparametric sign test arises as a natural limiting procedure in this setting, and the technical development of this article shows how optimality properties derived by Huber~\cite{Huber} can be applied in the semi-supervised setting to ensure that the maximal model selection error induced by automatic labeling is minimized.  Thusly one arrives at an algorithmic procedure that compares the relative performance of two competing systems in order to test the significance of performance differences between them, and hence to select the model that is ``closest'' (in the sense of Kullback-Leibler divergence) to the true data-generating distribution.

\subsection{Unlabeled Data in the Context of Speech Processing}

To clarify the notions of supervised/semi-supervised learning and
labeled/unlabeled data in the speech processing context at hand, we briefly
recall the standard machine learning paradigm as follows.
Fundamentally, one assumes the existence of an unknown joint
probability distribution $p_{X,Y}(x,y) \neq p_{X}(x) p_{Y}(y)$, from
which a number of independent and identically distributed samples
$(x_1,y_1), (x_2,y_2), \ldots$ are available; these are termed
\emph{training data}, and are used to fit a model that predicts values
taken by $Y$ based on observed instances of $X$.   In classification
tasks $Y$ is a discrete random variable, and its range of possible
values comprises the set of \emph{labels}---corresponding to, for
example, an orthographic transcript of the word or phrase represented
by an instance of acoustic waveform data $X$.

The goal in a traditional \emph{supervised learning} scenario is to
devise algorithms that strike a balance between fidelity to the set of
labeled examples $\{(x_i,y_i)\}$, and effective generalization to
other as-of-yet unseen \emph{test data} comprising additional
observations of $X$---a classical bias-variance trade-off between
model goodness-of-fit and generalization properties.  This trade-off is typically optimized by calculating empirical error rates on an additional ``held-out'' set
of labeled data for which ground truth is known, in a manner similar
to parameter estimation via cross-validation.

Fitting a model to accomplish this goal is thus mathematically
equivalent to building a system, and one speaks of the ``training'' or
model-building stage, and the ``testing'' or application stage, in
which a system is subsequently deployed and put into practical
use---and which assumes that both training and test data are drawn
from the same probability distribution.  When this assumption is
satisfied, it is clear that speech engineering systems benefit
directly from ever-greater amounts of \emph{labeled training data}.
Time, money, and expertise, however, typically limit the amount of
such data available in any given application scenario of interest.
 It is thus of much interest to develop algorithms that are built
using some amount of labeled training data, but whose performance can
be further improved through careful use of \emph{unlabeled} data---the
so-called semi-supervised learning paradigm \cite{Chapelle}.

Thus far, the application of semi-supervised methods to speech
processing has been limited to ideas such as data augmentation \cite{WessellN} or self-training~\cite{ma}, each of which involves re-fitting
the models under consideration---and hence rebuilding the
corresponding speech engineering systems.  While such approaches have
shown promise, such extreme re-fitting may not be desirable---or even
possible---in certain settings, for instance when a large-scale
system is already deployed and must be adapted to new test conditions.

Speech engineering is thus ripe for the introduction of new
semi-supervised learning approaches; not only can nearly limitless amounts of
acoustic waveform data be acquired from a variety of digital sources,
but also many algorithms have matured to the point that performance
improvements are often driven simply by increasing the amount of labeled
training data.  Employing unlabeled data to directly improve existing
approaches, however, \emph{requires inferring the labels}---and in
this context, a natural but unsolved problem is to understand whether
and how \emph{automatically labeled data} taken as output from current
systems can be used to this effect.  As indicated above, this article brings ideas from robust statistics and likelihood-based model selection to bear on this problem, and introduces not only a framework to analyze the errors resulting from automatic labeling, but also a practical means of treating them.

The article is organized as follows.  Section \ref{sec:lr} develops
likelihood-based semi-supervised model selection techniques, first
considering the case of labeled data, and subsequently the unlabeled case.  Section
\ref{sec:unpron} then formulates this semi-supervised framework in the
speech processing context of selecting from amongst competing
pronunciation models to optimize system performance.  Large-scale
experiments with well-known data sets in Section \ref{sec:large} then
demonstrate that this approach achieves state-of-the-art performance in the context of speech recognition, spoken term detection, and phonemic similarity to a given reference, even when compared to the conventional \emph{supervised} method of forced alignments to reference
orthographic transcripts.  Section~\ref{sec:dis} concludes the article
with a discussion of these results and their implication for improving
speech processing through the use of unlabeled development data.

\section{Theory: Likelihood-Based Model Selection}
\label{sec:lr}

Viewed from a machine learning perspective, parametric statistical
models are directly instantiated as large-scale speech processing
systems.  Labeled data are used to fit model parameters in the manner
described above; e.g., to estimate the state transition matrix
of a hidden Markov model.  In addition, one must also typically fit a modest
number of parameters that alter the structure or function of the model class under consideration;
for instance, in automatic speech recognition, the marginal acoustic
likelihood of an utterance typically depends on a model for the
pronunciation(s) of a given word---a setting we return to in Section
\ref{sec:unpron}.

When training and test conditions match exactly, all parameters can be
fitted simultaneously during the training stage, using principled and
efficient procedures such as the expectation-maximization algorithm.
In practice, however, it may be the case that only a small amount of
labeled training data is well matched to the conditions that prevail
during test---precluding even cross-validation as an option---or
that a deployed system must be adapted to new test conditions in the
absence of its original training data.  In such cases it is typical to
set aside a small amount of \emph{development data} for purposes of
\emph{model selection} as follows.

\subsection{The Supervised Case: Labeled Development Data}
\label{sec:labDat}

Recall that in our setting, $X$ represents acoustic waveform data, and
hence is a continuous random variable.  The true but unknown
data-generating model, then, takes the form of a conditional
probability density function $p(x\,\vert\,y) : =
p_{X\,\vert\,Y}(x\,\vert\,Y=y)$.  When interpreted for fixed $X$ as a
function of unknown label $Y$, this density thus evaluates to the acoustic likelihood of $X$ for any given candidate label $Y = y$.

In practice, we have access to $p(x\,\vert\,y)$ only through the given
pairs of training samples $(x_1,y_1), (x_2,y_2), \ldots$, and we must
proceed in the absence of direct knowledge of the true
model.  Any speech processing system will in turn generate its own set of
putative acoustic likelihoods, and thus it is natural to
seek the likelihood function that is closest to the true data-generating
model $p(x\,\vert\,y)$, in hopes that this will yield the best
overall system performance.  This leads to a \emph{model selection}
problem in which we use the training samples at hand as a proxy for
$p(x\,\vert\,y)$, to choose amongst competing models and build a
system that can predict $Y$ given $X$ with minimal misclassification error.

Assume, then, that we have several competing sets of candidate models $p_1(x\,\vert\,y;\theta_1), p_2(x\,\vert\,y;\theta_2), \ldots$, each dependent on distinct parameter sets $\theta_1, \theta_2,
\ldots$, whose quality we wish to evaluate with respect to the true (but
unknown) model $p(x\,\vert\,y)$.  A natural approach is to evaluate
the Kullback-Leibler divergence of the ``best'' representative
$p_{k}(x\,\vert\,y;\theta_k^*)$ of each set from $p(x\,\vert\,y)$, with
$\theta_k^*$ the maximum-likelihood estimate of parameter set
$\theta_k$ as determined from the training data.  Thus we seek
\begin{align*}
\operatornamewithlimits{argmin}_{k} &
\operatorname{\mathbb{E}}_p \left( \log p(x\,\vert\,y) \right) -
\operatorname{\mathbb{E}}_p \left( \log p_{k}(x\,\vert\,y;\theta_k^*) \right)
\\ \equiv \operatornamewithlimits{argmax}_k &
\operatorname{\mathbb{E}}_p \left( \log p_{k}(x\,\vert\,y;\theta_k^*)
\right)
\text{,}
\end{align*}
with 
$-\operatorname{\mathbb{E}}_p \left( \log
p_{k}(x\,\vert\,y;\theta_k^*) \right)$ sometimes referred to as the
\emph{cross-entropy} of $p_k$ relative to $p$, and the corresponding
optimization task one of \emph{cross-entropy minimization}.

Under the assumption of independent and identically distributed pairs
of training examples, we may form an empirical estimate of each
cross-entropy simply by evaluating the respective data log-likelihoods
$\log p_{k}(x\,\vert\,y;\theta_k^*)$ with respect to each pair of
training samples, and forming the corresponding arithmetic averages.
Assuming the necessary technical conditions of \cite{Vuong}, it then
follows that we may formulate a multi-way hypothesis test amongst
models $p_1, p_2, \ldots$.  We later consider this multi-way setting in detail; however, for clarity of exposition, we first consider the case of only \emph{two} competing models $p_1$ and $p_2$, which admits
three possible outcomes:
\begin{align*}
\mathcal{H}_0: \operatorname{\mathbb{E}}_p \left( \log
p_{1}(x\,\vert\,y;\theta_1^*) \right)
& = \operatorname{\mathbb{E}}_p \left( \log
p_{2}(x\,\vert\,y;\theta_2^*) \right)
\\ \mathcal{H}_1: \operatorname{\mathbb{E}}_p \left( \log
p_{1}(x\,\vert\,y;\theta_1^*) \right)
& > \operatorname{\mathbb{E}}_p \left( \log
p_{2}(x\,\vert\,y;\theta_2^*) \right)
\\ \mathcal{H}_2: \operatorname{\mathbb{E}}_p \left( \log
p_{1}(x\,\vert\,y;\theta_1^*) \right)
& < \operatorname{\mathbb{E}}_p \left( \log
p_{2}(x\,\vert\,y;\theta_2^*) \right)
\text{.}
\end{align*}
Hypothesis $\mathcal{H}_k$ thus favors the $k$th competing model, with
the null hypothesis $\mathcal{H}_0$ representing their equivalence.

The natural test statistic in this \emph{labeled data} setting is then
given by the log-ratio of likelihoods $p_1, p_2$ described above,
evaluated with respect to training data---possibly even the same training data used to fit the
maximum-likelihood model parameter estimates $\theta_k^*$---as follows:
\begin{equation}\label{eq:Tlab}
T_{\mathrm{lab}} : =
\sum_j \log \frac{p_{1}(x_j\,\vert\,y_j;\theta_1^*)}{p_{2}(x_j\,\vert\,y_j;\theta_2^*)}
\text{.}
\end{equation}

The careful reader will note that in such a regime, where expectations
are defined with respect to some unknown distribution $p$, we are in
fact working with potentially \emph{misspecified} models $p_1$ and
$p_2$; see \cite{white82,kent} for properties of maximum-likelihood
estimation of the parameter sets $\theta_1$ and $\theta_2$ in this
setting; for our purposes it suffices to note that such estimators
still possess the requisite technical properties.

In the case of interest to us here, the conditional models $p_1$ and
$p_2$ are assumed to be \emph{strictly non-nested}, such that no
conditional distribution in $X$ given $Y$ can be achieved by both
$p_1$ and $p_2$.  Vuong \cite{Vuong} shows a central limit theorem for
this setting when $\mathcal{H}_0$ is in force, in that as the
number of training samples grows large, an appropriately standardized
version of the test statistic $T_{\mathrm{lab}}$ is asymptotically
distributed as a unit Normal.  (It is
straightforward to proceed in the absence of this assumption, with
appropriate adjustments to test statistic asymptotics.)  The necessary
normalization is given by the sample standard deviation of
log-likelihood ratio evaluations times the root of the number of
training samples; if $\mathcal{H}_0$ fails to be in force, then the value of this statistic
diverges (almost surely) to $\pm \infty$.

This result in turn implies a concrete directional test
for model selection: fixing a significance level $\alpha$ yields a
corresponding critical value $z_{\alpha/2}$ according to the standard
Normal distribution.  If the normalized test statistic evaluates to
greater than $z_{\alpha/2}$, we select model $p_1$; if it evaluates to
less than $-z_{\alpha/2}$, we decide in favor of model $p_2$.  Otherwise,
we conclude that there is insufficient evidence to reject the
hypothesis $\mathcal{H}_0$ of model equivalence, and we conclude that
models $p_1$ and $p_2$ cannot be distinguished on the basis of the given
training data and chosen significance level.

\subsection{The Semi-Supervised Case: Unlabeled Development Data}
\label{subsec:clr}

Now suppose that our two competing models $p_1$ and $p_2$ have already been
``trained,'' such that $\theta_1, \theta_2$ have been fitted by
maximum-likelihood estimation to obtain $\theta_1^*, \theta_2^*$, but
that we wish to leverage $n$ additional \emph{unlabeled data} examples
$x_1, x_2, \ldots, x_n$ to accomplish the model selection task
described in Section~\ref{sec:labDat} above.  Lacking the
corresponding class labels $y_1, y_2, \ldots, y_n$ for these data, we
thus seek to employ \emph{automatically generated} labels $\hat{y}_1,
\hat{y}_2, \ldots, \hat{y}_n$ fitted respectively by
maximum-likelihood under each of the two systems, such that we replace
the conditional log-likelihood ratio of~\eqref{eq:Tlab} by the
generalized log-likelihood ratio
\begin{equation}\label{eq:Thatlab}
\sum_{i=1}^n \log
\frac{p_{1}(x_i\,\vert\,\hat{y}_i;\theta_1^*)}{p_{2}(x_i\,\vert\,\hat{y}_i;\theta_2^*)}
 = \sum_{i=1}^n \log \frac{\max_y
p_{1}(x_i\,\vert\,y;\theta_1^*)}{\max_y
p_{2}(x_i\,\vert\,y;\theta_2^*)}
\text{.}
\end{equation}

Of course, maximum-likelihood labeling (``decoding'') of $Y$ given $X$
incurs some error, and hence it is natural to ask under what
conditions we can replace $T_{\mathrm{lab}}$ in the labeled-data model
selection task of Section~\ref{sec:labDat} with~\eqref{eq:Thatlab}.
Since this 
corresponds to the use of labels \emph{taken as output from trained
systems}---i.e., estimated under each of the two competing models
$p_1$ and $p_2$---this procedure will inevitably suffer from
misclassification errors with respect to the estimated labels; if
systems $p_1$ and $p_2$ exhibit reasonable performance, however, the
corresponding marginal error rate $\epsilon$ will be small.  In the
limit as $\epsilon$ tends to zero, of course, we recover precisely the setting of labeled data encountered in
Section~\ref{sec:labDat} above.

For the case of small but nonzero $\epsilon$, and assuming now that the
true data-generating model is either $p_1$ or $p_2$, we show below
that a principled model selection procedure may obtained by adapting results from
the labeled-data setting as follows.  Each individual likelihood
ratio $p_{1}(x_i\,\vert\,\hat{y}_i;\theta_1^*) /
p_{2}(x_i\,\vert\,\hat{y}_i;\theta_2^*)$ will instead be \emph{censored}, by
bounding its range from above and below in
order to limit the influence of misclassification errors on the
overall model selection procedure.  In the limit, as we will see, this
recovers the well-known nonparametric \emph{sign test}, which simply
tabulates for every $i = 1, 2, \ldots, n$ the sign of each log-likelihood ratio, rather than its actual value. As we formulate in Section~\ref{sec:efficiency} below, this approach sacrifices a degree of statistical efficiency for enhanced robustness, which in turn enables the influence of errors in the set $\{\hat{y}_i\}$ of automatically generated labels to be limited.

Not only is this approach intuitively reasonable, but it is also
provably optimal in a minimax sense, as we now describe.  To account
for the misclassification errors induced by automatic labeling, we
model the consequence of this inexact labeling procedure by replacing
the exact conditional densities $p_1(x\,\vert\,y)$ and
$p_2(x\,\vert\,y)$ with \emph{mixtures} of these densities and
``contaminating'' distributions that represent the aggregate effects
of misclassification.  The misclassification error rate $\epsilon \ll
1$ moreover serves as the mixture weight for each respective
contaminating density---the so-called $\epsilon$-contaminated case \cite{Huber}.

Rather than seeking to determine these contaminating distributions
directly, it is natural to ask if there exists a
\emph{least favorable} case: a form of contamination that, for fixed
$\epsilon$, would serve to maximize the probability of selecting the
incorrect model $p_1$ or $p_2$.  The answer is affirmative:  Amongst
all possible contaminating densities, we are guaranteed that a
\emph{least favorable} pair exists whenever the likelihood ratio
$p_1(x\,\vert\,y) / p_2(x\,\vert\,y)$ is monotone and $\epsilon$ is
small enough to ensure that the corresponding sets of
admissible $\epsilon$-contaminated mixtures remain disjoint.

In this case, a result obtained by Huber \cite[Theorem 3.2]{Huber} in
the context of robust statistics may be applied to show that, to
minimize this maximal risk of an error in model selection, it suffices
to consider a specific form of contamination of $p_1$ by $p_2$, and
vice-versa.  The precise mixture form required by Huber's result is
obtained by partitioning the range space of $p_1$ and $p_2$ in a
manner that depends on $b > a > 0$ as follows:
\begin{align*}
\tilde{p}_1(x\,\vert\,\cdot) & = (1-\epsilon) \cdot
\begin{cases}
 p_1(x\,\vert\,\cdot) & \text{whenever $p_1 > a p_2$,} \\
 a p_2(x\,\vert\,\cdot) & \text{otherwise;}
 \end{cases}
\\ \tilde{p}_2(x\,\vert\,\cdot) & = (1-\epsilon) \cdot
\begin{cases}
 p_2(x\,\vert\,\cdot) & \text{whenever $p_2 > b^{-1} p_1$,} \\
 b^{-1} p_1(x\,\vert\,\cdot) & \text{otherwise.}
 \end{cases}
\end{align*}
A likelihood ratio test based on $\tilde{p}_1 / \tilde{p}_2$ is thus
seen to yield
$$
\frac{\tilde{p}_1(x\,\vert\,\cdot)}{\tilde{p}_2(x\,\vert\,\cdot)} =
\begin{cases}
 a & \text{if $p_1 / p_2 \leq a$,} \\
 p_1(x\,\vert\,\cdot) / p_2(x\,\vert\,\cdot) & \text{if $a < p_1 / p_2 < b$,} \\
 b & \text{if $p_1 / p_2 \geq b$,}
 \end{cases}
$$
and hence we have arrived at the minimax test for the case of
$\epsilon$-contaminated densities $p_1$ and $p_2$---a test based on
likelihood ratio evaluations censored from below at $a$ and above at $b$.

As noted by Huber, the limiting case occurs when
$\epsilon$ is sufficiently large that the sets of
$\epsilon$-contaminated mixture densities $\tilde{p}_1 , \tilde{p}_2$
cease to be disjoint, and begin to overlap; in our setting, this corresponds to
the limit as $a$ and $b$ both approach unity. As $a$ and $b$ both approach unity, the log-likelihood ratio reflects only which term of the comparison is larger, yielding the
\emph{sign test} for model selection as described above:
\begin{equation}\label{eq:Tunlab}
T_{\mathrm{unlab}} : =
\# \left\{i: \log
\frac{p_{1}(x_i\,\vert\,\hat{y}_i;\theta_1^*)}{p_{2}(x_i\,\vert\,\hat{y}_i;\theta_2^*)}
>0 \right\}
\text{.}
\end{equation}
This test statistic is distributed as a sum of $n$ Bernoulli trials
whenever the unlabeled examples $x_1,x_2,\ldots,x_n$ are independent
and identically distributed, and is hence a binomial
random variable.  As such, we obtain a concrete directional test for
model selection in the semi-supervised setting, in a manner that generalizes the
supervised setting of Section~\ref{sec:labDat} above.

As in the supervised case, we may fix a significance level $\alpha$ and determine a corresponding critical value $k_{\alpha}$ according to the binomial distribution with parameters $n$ and $p$, where $p = 1/2$ under the null hypothesis of model equivalence.  For a one-sided upper-tail test of size $\alpha$, we reject $\mathcal{H}_0$ in favor of $\mathcal{H}_1$ if $T_{\mathrm{unlab}} > k_{\alpha}$, where $k_{\alpha}$ is the smallest integer such that
$\sum_{k=k_{\alpha}}^{n}{n\choose k}\left(\textstyle\frac{1}{2}\right)^{n} \leq \alpha
\text{;}
$
reversing this inequality and summing from zero to $k_{\alpha}$ yields the corresponding one-sided lower-tail test.  For a fixed alternate with $p \neq 1/2$, the corresponding probability of correct selection is given by
$\sum_{k=k_{\alpha}+1}^n {n\choose k} p^k(1-p)^{n-k}
\text{.}
$
The sign test has many appealing properties; we next investigate its statistical efficiency in this context, and refer the reader to \cite{Lehmann} for other results.

\subsection{Analysis: Comparing Statistical Efficacy and Efficiency}
\label{sec:efficiency}

To summarize the results of  \cite{Huber} and \cite{Vuong} as they apply to our discussion of model selection above, the best test in the case of \emph{labeled} development data accumulates the log-likelihood ratios of each example $x_i$ given its correct label $y_i$, while in the case of \emph{unlabeled} development data the corresponding minimax test accumulates the \emph{signs} of these ratios when evaluated with respect to each automatically generated label $\hat{y}_i$.  To compare the statistical efficacy of these two testing procedures, we may compute their \emph{asymptotic relative efficiency} under general assumptions regarding the limiting distributions of (suitably standardized versions of) test statistics $T_{\mathrm{lab}}$ of~\eqref{eq:Tlab} and $T_{\mathrm{unlab}}$ of~\eqref{eq:Tunlab} obtained under the null hypothesis.

Asymptotic relative efficiency expresses the limiting ratio of sample sizes necessary for two respective tests to achieve the same power and level against a common alternative; if one test has an asymptotic efficiency of 50\% relative to another, then the former requires twice as many samples (in the large-sample limit) to achieve the same performance.  Its computation requires knowledge of the asymptotic distributions of both test statistics under the null hypothesis, as we now describe.

Recall that when comparing strictly non-nested models using labeled data, a limit theorem holds under the null; let $f(\cdot)$ denote the associated density function, with corresponding variance $\sigma^2$.  The so-called \emph{efficacy} of the labeled-data test is in turn given by $1/\sigma$ under suitable regularity conditions, with that of the unlabeled-data sign test given by $2f(0)$ when $T_{\mathrm{unlab}}$ is appropriately standardized \cite{Lehmann}.

The corresponding asymptotic relative efficiency is in turn given by the squared ratio of test efficacies, which evaluates to the quantity $[2\sigma f(0)]^2$.  This result implies that when $T_{\mathrm{lab}}$ is asymptotically Normal, the sign test corresponding to~\eqref{eq:Tunlab} is only $2/\pi \approxeq 64\%$ as efficient as the labeled-data test corresponding to~\eqref{eq:Tlab}, since $(2\sigma/\sqrt{2\pi \sigma^2})^2 = 2/\pi$.  We may in fact generalize this result slightly by following the analysis of \cite{Kanefsky}, and considering the so-called generalized Gaussian distribution with location parameter $\mu$ and scale parameter $\sigma$:
\begin{equation*}
f_p(x) = \frac{1}{2\sigma \zeta(p)^{1/p} \, \Gamma(1+1/p)}
\exp \! \left[- { \textstyle \frac{1}{\zeta(p)} }
\left(\frac{|x-\mu|}{\sigma}\right)^{\!p} \, \right]
\text{.}
\end{equation*}
Here $\Gamma(\cdot)$ is the Gamma function, $\zeta(p) = [\Gamma(1/p) / \Gamma(3/p) ]^{p/2}$, and exponent $1 \leq p \leq 2$ allows us to interpolate between the Laplacian ($p=1$) and Normal ($p=2$) densities.

\begin{figure}[t]
  \centering
   \includegraphics[width=\columnwidth]{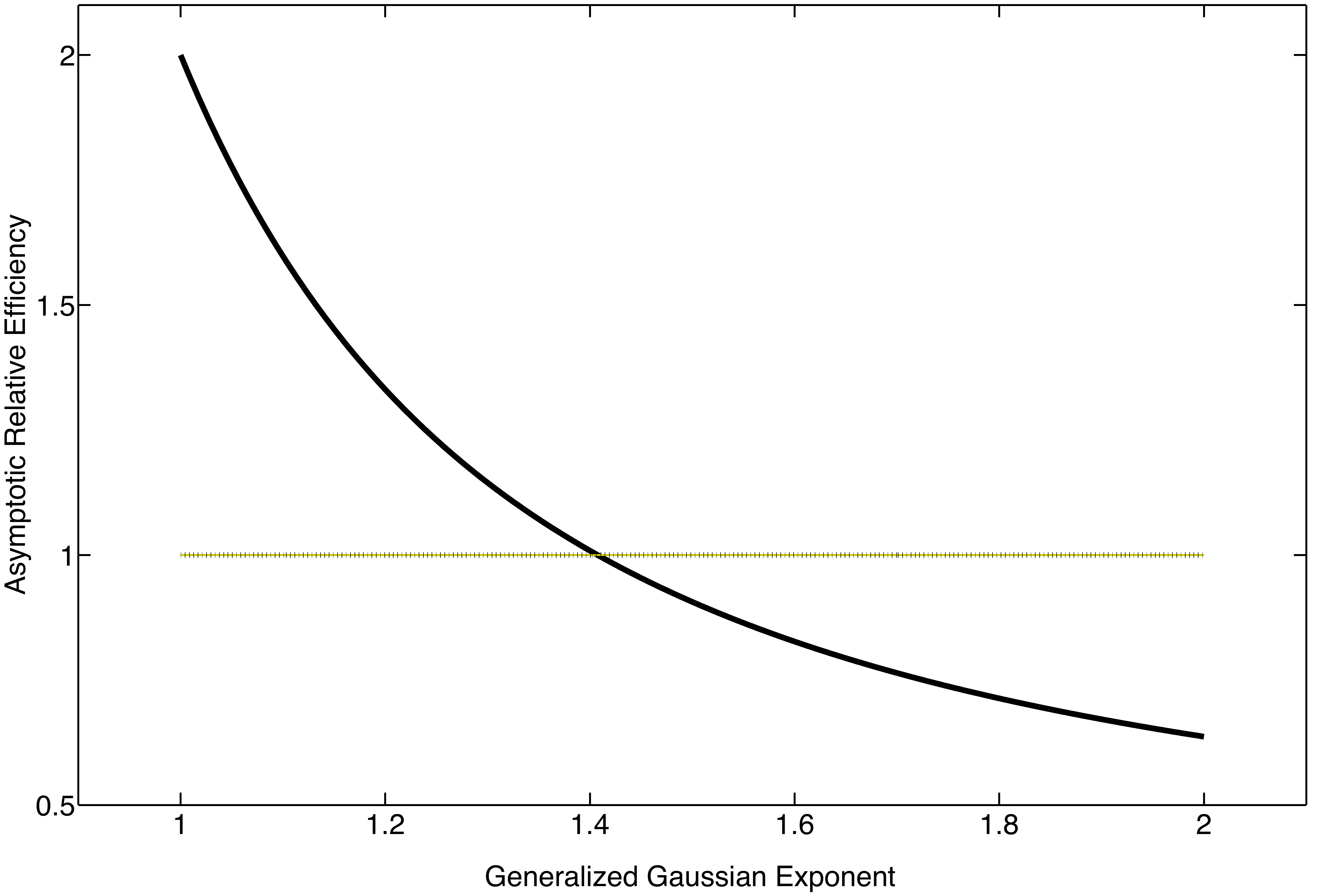}
\caption{Asymptotic relative efficiency of tests in the semi-supervised versus supervised settings, when the test statistic of the latter converges to a generalized Gaussian distribution with exponent $p$ between 1 (Laplacian) and 2 (Normal).  The horizontal line divides the range of $p$ into cases for which the sign test is less efficient than the conventional likelihood ratio test, as in the case of the Normal, and vice-versa.}
\label{fig:are}
\end{figure}

If we thus consider the expression $[2\sigma f_p(0)]^2$ for asymptotic relative efficiency, it follows from the relation $\Gamma(1+1/p) = \Gamma(1/p)/p$ that, as a function of exponent $p \in [1,2]$, the asymptotic relative efficiency for the case of a generalized Gaussian distribution having exponent $p$ is $p^2 \Gamma(3/p) / [\Gamma(1/p)]^3$.  This result is illustrated in Figure \ref{fig:are},
which confirms that, were the asymptotic distribution of $T_{\mathrm{lab}}$ to approach a Laplacian density with $p=1$, rather than a Normal with $p=2$, the sign test would be twice as efficient in the large-sample limit.

\subsection{Selecting from Amongst $k>2$ Competing Models}
\label{sec:multiway}

As demonstrated above, the case of two competing hypotheses yields theoretical performance guarantees; however, in practice it is often necessary to select from amongst $k>2$ models.  While optimality is no longer necessarily retained~\cite{Huber}, this problem is of sufficient practical interest to have generated a large contemporary literature in machine learning~\cite{ana,allwein}. Ê

Of the many approaches described in, e.g.,~\cite{ana,allwein}, several feature pairwise comparisons: in the so-called ``one vs.~all'' method, each model is assigned a real-valued score relative to all others, and the model with the highest overall score is selected.  Other possibly approaches include ``tournament-style,'' following initial pairwise comparisons, or the case of all possible $\binom{k}{2}$ pairwise comparisons.  

The latter approach has been suggested in~\cite{rhyne} for the case of the sign test, and currently remains common practice within the machine learning community, despite multi-class procedures tailored to specific learning methods \cite{ana}.  As such, we employ it to select amongst competing pronunciation models in our experiments below.

\section{Application: Selecting Pronunciation Models}
\label{sec:unpron}

As a prototype application of the semi-supervised model selection approach derived in Section~\ref{sec:lr}, we now consider the task of evaluating candidate pronunciations of spoken words in large-scale speech processing tasks.  To select amongst competing pronunciations, we consider two speech recognition systems that differ only in the pronunciation of a particular word, and show how to employ both the conventional test of~\eqref{eq:Tlab} using \emph{transcribed} audio data, and the sign test of~\eqref{eq:Tunlab} using \emph{untranscribed} audio data.

\subsection{Motivation for Semi-Supervised Pronunciation Selection}

The selection of pronunciation models is crucial to several speech processing applications, including large-vocabulary continuous speech recognition, spoken term detection, and speech synthesis, each of which requires knowledge of the pronunciation(s) of each word of interest.  In this setting, a set of admissible pronunciations forms what is termed a \emph{pronunciation lexicon}, which comprises mappings from an orthographic form of a given word (e.g., {\it tornados}) to a phonetic form (e.g., /t er n ey d ow z/).

The conventional means of creating a pronunciation lexicon is to employ a trained linguist. However, as is the case with other examples requiring data to be hand-labeled by experts, this process is expensive, inconsistent, and even at times impossible, when individuals lack sufficiently broad expertise to create pronunciations for all words of interest \cite{riley}.  In turn, several approaches for automatically \emph{generating} pronunciations have been put forward \cite{riley, lucassen, vitale, ramabhadran, beaufays, teppermann}, and inevitably a model selection decision must be made to choose between candidate pronunciations.  However, these approaches have themselves relied upon labeled training data, in the form of spoken examples of a given word and the corresponding orthographic transcripts.

In addition to the initial creation of a lexicon, pronunciation models are also necessary to maintain the vocabulary of speech processing systems over time:  Although the pronunciation lexicon for a given system is created for as large a vocabulary as possible before deployment, this lexicon must be extended over time to incorporate \emph{out-of-vocabulary} words.  Such terms can be new words or names that come into common usage, rare or foreign words, or simply words not deemed significantly important at the time a system's lexicon was constructed.  Dynamically adjusting to changing vocabularies thus requires the generation of \emph{new} pronunciations over time, thereby reinforcing the need for an efficient and effective means of automatically selecting from amongst candidate pronunciations \cite{mamou, white2, white1}.

\subsection{Methods for Selecting a Pronunciation Model}

Much effort to date has been focused in the area of automatic pronunciation modeling---i.e., grapheme-to-phoneme or letter-to-sound rules.  Previous work, including \cite{riley} and \cite{lucassen}, has attempted to simultaneously generate a set of pronunciations and select between them. Also, work including \cite{deng2} augments the possible pronunciations by building a larger phone network to select the pronunciation. Additional resources are typically required, including existing pronunciation lexica \cite{riley}, speech samples \cite{ramabhadran, beaufays}, linguistic rules \cite{teppermann}, or a combination of these. The focus of previous work has been on pronunciation variation \cite{riley, ramabhadran} or on common words \cite{lucassen, vitale}. Note that in practice, other concerns may dictate choices between competing pronunciations, such as the scenario considered in \cite{deng1}, while highlighting the trade-offs between word accuracy and overall word error rate (WER). In the current setting, however, we are agnostic as to how the pronunciations are generated; our goal is simply to choose between them. 

To this end, consider the setting in which we have example utterances $\{x_1, x_2, \ldots, x_n\}$, their corresponding transcripts $\{y_i\}$, and two ``trained'' speech recognition systems $p_1(\cdot)$ and $p_2(\cdot)$ that are identical (i.e., conditioned on the same parameters) \emph{except} that for one word, models $p_1$ and $p_2$ use different pronunciations, say $\theta_1^*$ for $p_1(\cdot;\theta_1^*)$ and $\theta_2^*$ for $p_2(\cdot;\theta_2^*)$.  This corresponds to the case of \emph{strictly non-nested} models outlined in Section~\ref{sec:lr}.  We subsequently describe and compare a supervised and semi-supervised method to select between candidate pronunciations $\theta_1^*$ and $\theta_2^*$, and hence between models $p_1$ and $p_2$, in settings where candidate words are analyzed one at a time (as opposed to comparing entire pronunciation lexicons).

\subsubsection{Supervised Selection of Pronunciations}
\label{sec:exp2}

The conventional mechanism for choosing between reference pronunciations of a word, examples of which are shown in Table~\ref{tab:hyp},
 \begin{table}
\center
 \begin{tabular}{ccccc}
	     Word & Candidate Pron. & Reference Pron. \\
		{\it guerilla} & g ax r ax l ax & g ax r ih l ax\\
		{\it guerilla} & {\bf g w eh r ih l ax} &\\
		{\it tornados} & {\bf t er n ey d ow z} & t er n ey d ow z\\
		{\it tornados} & t ao r n ey d ow s  & t ow r n ey d ow z\\
\end{tabular}
\caption{Examples of candidate and reference pronunciations \label{tab:hyp}}
\end{table}
is to acquire spoken utterances that contain the word, along with an orthographic transcription of the utterances, and compute a forced alignment of the acoustic waveform data to the transcripts, first using one pronunciation and then using the other \cite{riley, lucassen, beaufays, teppermann}. The pronunciation that is assigned a higher (Viterbi maximum likelihood) score during alignment is then chosen. For each word there are a fixed number of candidate pronunciations, with at least one (e.g., {\it guerilla}) reference pronunciation per word, although there may be several (e.g., {\it tornados}).

Cast in the notation of Section~\ref{sec:lr}, the conventional \emph{supervised} method of pronunciation selection proceeds as follows:

\begin{enumerate}
\item Use the sequence of words comprising reference transcription $\bm{y}_i^{(\mathrm{ref})}$ for utterance $X_i=x_i$ to compute the log-likelihood ratio
\begin{align*}
\Lambda_i(x_i|\theta_1^*,\theta_2^*,\bm{y}_i^{(\mathrm{ref})}) = & \sum_{y \in \bm{y}_i^{(\mathrm{ref})}} \log  p_1(x_i|y;\theta_1^*) \\
 - & \sum_{y \in \bm{y}_i^{(\mathrm{ref})}} \log p_2(x_i|y;\theta_2^*)
 \text{;}
\end{align*}
\item Use the $n$ utterances to form $T_{\mathrm{lab}}$ and test as follows:
\begin{equation}
\label{eq:sup2}
T_{\mathrm{lab}} = \sum_{i=1}^n \Lambda_i(x_i|\theta_1^*,\theta_2^*,\bm{y}_i^{(\mathrm{ref})})  \quad {{\mathcal{H}_{1} \atop >} \atop {< \atop \mathcal{H}_{2}}} \quad \tau_{\mathrm{lab}} \text{;}
\end{equation}
\item Decide between $\mathcal{H}_{1}$ (model/pronunciation $\theta_1^*$) and $\mathcal{H}_{2}$ (model/pronunciation $\theta_2^*$)  based on the difference in conditional likelihood evaluations, given forced-alignment reference transcripts, as indicated in~\eqref{eq:sup2}.
\end{enumerate}

\subsubsection{Semi-Supervised Pronunciation Selection}
\label{sec:expun}

The conventional method of pronunciation selection described above requires \emph{transcribed audio data} whose production is a difficult, time-consuming, and laborious task.  In many applications, external information can potentially alleviate the need for transcriptions by identifying recorded speech segments that are a priori likely to contain instances of a given word, which in turn may be used to select between candidate pronunciations.  Examples include news items and television shows, each of which provides a rich source of \emph{untranscribed} speech that could serve to improve the selection of pronunciations.

It is furthermore often the case that, while a transcript corresponding to spoken examples of a word is unavailable, we may have some knowledge that it has occurred in a particular audio archive.  For example, we may know from weather records that a broadcast news episode recently aired about natural disasters, giving us a degree of confidence that instances of words like {\it tornados} are likely to appear.  We may not know where or how many times such a word occurs in a particular audio segment, but we can still use the entire broadcast to help us choose between candidate pronunciations for {\it tornados}, examples of which are given in
Table \ref{tab:hyp}.

In the {\it absence} of labeled examples we proposed to use the recognition system outputs themselves---unconstrained by any forced alignment or reference transcript---to select between candidate pronunciations.  Each speech recognition system is run on every candidate data segment likely to contain a given word of interest, and from these results the corresponding acoustic likelihoods are evaluated with respect to the entire data set, leading to the selection of the candidate pronunciation yielding the highest overall likelihood.

Recalling our notation for the competing models $p_1(\cdot;\theta_1^*)$ and $p_2(\cdot;\theta_2^*)$, with corresponding pronunciations $\theta_1^*$ and $\theta_2^*$, this semi-supervised approach proceeds in analogy to the labeled-data setting as follows:
\begin{enumerate}

\item Form the automatically generated word sequences $\bm{{\hat{y}}}_i^{(\theta_1^*)}$ and $\bm{{\hat{y}}}_i^{(\theta_2^*)}$ for each utterance $X_i=x_i$:
\begin{align*}
\bm{{\hat{y}}}_i^{(\theta_1^*)} & = \operatornamewithlimits{argmax}_{y} p_1(y|{x_i};\theta_1^*)\\
\bm{{\hat{y}}}_i^{(\theta_2^*)} & = \operatornamewithlimits{argmax}_{y} p_2(y|{x_i};\theta_2^*),
\end{align*}
and use $\bm{{\hat{y}}}_i^{(\theta_1^*)},\bm{{\hat{y}}}_i^{(\theta_2^*)}$ to compute the log-likelihood ratio
\begin{align*}
\Lambda_i(x_i|\theta_1^*,\theta_2^*) =& \sum_{y \in \bm{{\hat{y}}}_i^{(\theta_1^*)}} \log  p_1(x_i|y;\theta_1^*) \\
-& \sum_{y \in \bm{{\hat{y}}}_i^{(\theta_2^*)}} \log p_2(x_i|y;\theta_2^*)
\text{;}
\end{align*}

\item Use the $n$ utterances to form $T_{\mathrm{unlab}}$ and test as follows:
\begin{equation}
\label{eq:unsup22}
T_{\mathrm{unlab}} =
\# \left\{i: \Lambda_i(x_i|\theta_1^*,\theta_2^*) > 0 \right\}
\,\,\, {{\mathcal{H}_{1} \atop >} \atop {< \atop \mathcal{H}_{2}}} \,\,\, \tau_{\mathrm{unlab}}
\text{;}
\end{equation}

\item Decide between $\mathcal{H}_{1}$ (model/pronunciation $\theta_1^*$) and $\mathcal{H}_{2}$ (model/pronunciation $\theta_2^*$) based on the number of log-likelihood ratios that evaluate to be positive, as indicated in~\eqref{eq:unsup22}.
\end{enumerate}

\section{Large-Scale Experimental Validation}
\label{sec:large}

We now present an experimental validation of the semi-supervised model selection approach presented in the preceding sections, consisting of selecting between candidate pronunciations in the context of three prototypical large-scale speech processing tasks.  For each of 500 different words, forced alignment and recognition outputs were produced for every pair of pronunciation candidates.  Recognition was performed on an hour of speech for every word and each corresponding candidate, making sure to include somewhere in the data to be recognized the same speech utterances that were used in the forced-alignment setting, yielding a total of 1000 hours of recognized speech.

The quality of the selected pronunciations was then evaluated in three different ways: through decision-error trade-off curves for spoken term detection, phone error rates relative to a hand-crafted pronunciation lexicon, and word error rates for large-vocabulary continuous speech recognition.  All experiments were conducted using well-known data sets, and state-of-the-art recognition, indexing, and retrieval systems.

\subsection{Methods and Data}
\label{sec:exp}

In order to evaluate the performance of semi-supervised pronunciation selection and its suitability for a variety of applications (e.g., recognition, retrieval, synthesis), and for a variety of word types (e.g., names, places, rare/foreign words), we selected speech from an English-language broadcast news corpus and identified 500 single words of interest.  Common English words were removed from consideration, to ensure that words of interest would often be absent from lexicons, and thus would require pronunciation selection (e.g., {\it Natalie}, {\it Putin}, {\it Holloway}), and all words of interest featured in at least 5 acoustic instances.  The selected words of interest were verified to be absent from the recognition system's vocabulary, and all speech utterances containing these words were removed from consideration during the acoustic model training stage.

For each word of interest, two candidate pronunciations were considered, each of which was generated by one of two different letter-to-sound systems \cite{sethy09}; furthermore, the 500 chosen words all had the property that the two letter-to-sound systems produced different pronunciations for them. For all subsequent experiments in semi-supervised pronunciation model selection, the sign test threshold $\tau_{\mathrm{unlab}}$ was set at $\tau_{\mathrm{unlab}} = n/2 + 1$, so that if more than half of the log-likelihood ratios evaluated to be positive, then the corresponding pronunciation model was chosen (i.e., a ``winner-takes-all'' approach). The threshold reflects our a priori belief of equally likely candidates, while enforcing our practical goal that one candidate or the other must be selected. The sensitivity to the threshold depends on the ``distance'' between models, as well as the number of observations. For the experiments in supervised pronunciation model selection, the threshold $\tau_{\mathrm{lab}}$ was set at zero, so that the candidate with the higher log-likelihood was chosen. 

To accomplish these experiments, a large-vocabulary continuous speech recognition (LVCSR) system was built using the IBM Speech Recognition Toolkit \cite{IBM} with acoustic models trained on 300 hours of HUB4 data. Around 100 hours were used as the test set for recognition word error rate and spoken term detection experiments. The language model for the LVCSR system was trained on 400M words from various text sources. The LVCSR system's word error rate on a standard broadcast news test set RT04 (i.e., distinct from the 100 hours used for the test set employed below) was 19.4\%. This LVCSR system was also used for lattice generation in the spoken term detection task. The OpenFST-based Spoken Term Detection system described in \cite{saraclar} was used to index the lattices and search for the 500 words of interest.  For additional details regarding the experimental procedures and data sets, the reader is referred to \cite{white3}.

\subsection{Experimental Procedure}
\label{sec:ExpProc}

To summarize the experimental procedure, two alternative pronunciations are generated by two different letter-to-sound systems for each of a set of 500 selected words. We also have a {\it reference} pronunciation for these words from a hand-crafted pronunciation lexicon. We assume for the purposes of these experiments that the reference pronunciation is not available, and we set ourselves the task of choosing between two alternative pronunciations for each word, evaluated with respect to three different metrics, as will be discussed below.

The choice between the two pronunciations is made via either the supervised method of Section \ref{sec:exp2} (denoted \emph{sup}) or the semi-supervised method of Section \ref{sec:expun} (denoted \emph{semi-sup}):
\begin{itemize}
\item \emph{Sup} selects the candidate pronunciation based on supervised forced alignment with a reference transcript;
\item \emph{Semi-sup} selects the candidate pronunciation based on unconstrained (i.e., fully automatic) recognition.
\end{itemize}
Some example words of interest and their accumulated test statistics are shown in Table \ref{tab:pgaus}.
\begin{table}[t]
\center
 \begin{tabular}{lrrr}
            Word & No. Samples & $|T_{\mathrm{lab}}|$ & $|T_{\mathrm{unlab}}|$\\
            {\it Acela} & 8 & 151.92 & 4 \\
            {\it afterwards} & 38 & 4846.52 & 31 \\
            {\it Albright} & 247 & 34118.11 & 230 \\
            {\it Barone} & 16 & 3011.04 & 12 \\
            {\it Beatty} & 5 & 359.75 & 5 \\
            {\it Iverson} & 21 & 1698.90 & 18\\
            {\it Peltier} & 12 & 741.12 & 9\\
            {\it Villanova} & 6 & 902.04 & 3\\
\end{tabular}
\caption{Example words and their accumulated test statistics \label{tab:pgaus}}
\end{table}
For each word, the number of true speech samples is listed, along with the accumulated log-likelihood ratios in accordance with \eqref{eq:sup2}, and the corresponding number of accumulated sign-test samples as per \eqref{eq:unsup22}, in which the effect of likelihood censoring is apparent.

Additionally, we compare the methods described above with an {\it oracle} and an {\it anti-oracle}, defined with respect to the hand-crafted lexicon as follows:
\begin{itemize}
\item The {\it oracle} selects the candidate that has the {\it smallest} edit distance to a reference pronunciation of that word
\item The {\it anti-oracle} selects the candidate that has the {\it largest} edit distance to a reference pronunciation of that word
\end{itemize}
To illustrate this notion, recall the earlier examples featured in  Table \ref{tab:hyp}, which lists two words, each with two hypothesized pronunciations.  In the case of these examples, the {\it oracle} pronunciation selection method would select the entries `/g ax r ax l ax/' and `/t er n ey d ow z/'.

\subsection{Results}
\label{sec:res}

\subsubsection{Spoken Term Detection}
\label{ssec:subhead}

\begin{figure}[t]
 \centering
  \includegraphics[width=\columnwidth]{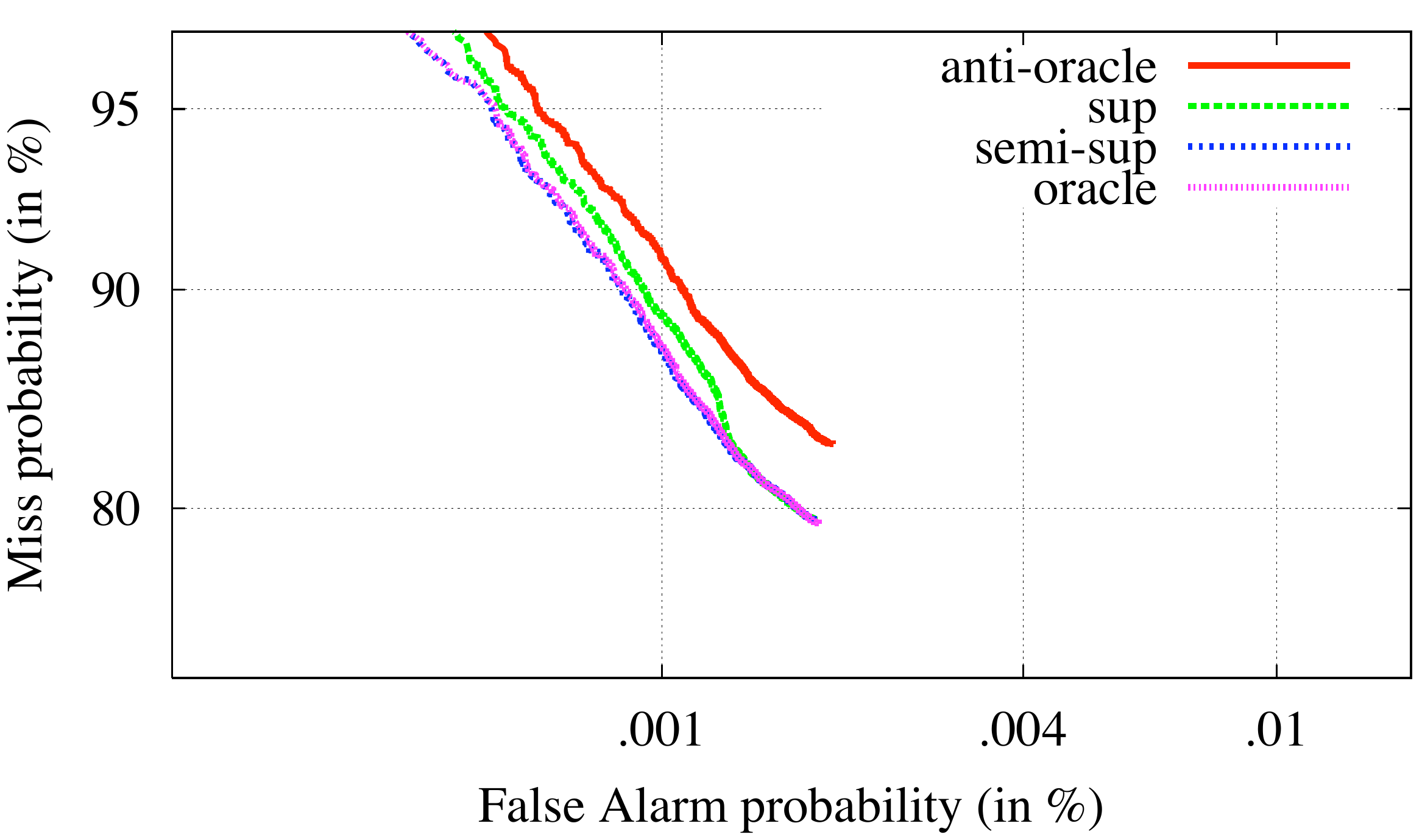}
 \caption{\label{fig:res}Decision-error trade-off curves for a spoken term detection task \cite{white3}, generated from 100 hours of speech data, using chosen pronunciations as queries to a phonetic/word-fragment index.  Note that {\it semi-sup} and {\it oracle} overlap at nearly all operating points.}
\end{figure}

Experimental results from \cite{white3}, showing the result of competing approaches to selecting between candidate pronunciations for purposes of spoken term detection, are shown in Fig.~\ref{fig:res}.
Lattices generated by the LVCSR system for the 100-hour test set were indexed and used for spoken term detection experiments in the OpenFST-based architecture described in \cite{saraclar}; the chosen pronunciations were used as queries to the spoken term detection system. Results from the OpenFST-based indexing system were computed using standard formulas from the National Institute of Standards and Technology (NIST) and scoring functions/tools from the NIST 2006 spoken term detection evaluation.  Note that the decision-error trade-off curves demonstrate that \emph{semi-sup} performs better than the supervised method for detection at nearly all operating points.

 \begin{table*}[t]
\center
\begin{tabular}{c|ccr|ccr|ccr}
	     {Method} &  {System Quality} & {No. Words} & PER\%  &  {System Quality} & {No. Words} & PER\%  &  {System Quality} & {No. Words} & PER\%\\
	     	 &  {(RT04 WER\%)} & {Resolved} & & {(RT04 WER\%)} & {Resolved} & &{(RT04 WER\%)} & {Resolved} & \\
  	     {\it sup}  & {29.3} & {359} & 13.00 & {24.5} & {390} & 13.66 & {19.4} & {449} & 14.50\\
  	     {\it semi-sup}  & {29.3} & {359} & 12.64 & {24.5} & {390} & 13.19 & {19.4} & {449} & 13.87\\
\end{tabular}
\caption{Phone error rates (PER) with respect to a hand-crafted lexicon \label{tab:per}}
\end{table*}

\subsubsection{Phone Error Rate (PER)}

This experiment measures which method---supervised or semi-supervised---selects pronunciations that have smaller edit distance to a reference pronunciation.  Referring again to Table \ref{tab:hyp} as an example, if the bolded pronunciations had been selected based on the observed speech data, there would be 2 errors out of 6 phones with respect to the closest reference pronunciation for {\it guerilla}: delete /w/ and change /er/ to /ax/, resulting in a 33\% PER; for {\it tornados}: 0\% PER.

We note that while the supervised method requires a few \emph{acoustic samples} of a word of interest, the semi-supervised method requires that a few instances of the word {\it be recognized}---correctly or incorrectly---by the LVCSR system. If insufficiently many instances are recognized, then a choice between alternative pronunciations cannot be made. Therefore, depending on the accuracy of the system, only a subset of the 500 words may be resolved (in the sense of having a pronunciation selected) by the semi-supervised method.  Consequently, we employed three different levels of language model pruning to yield three levels of system quality, defined in terms of word error rate on the standard RT04 data set.  The resultant error rates on the RT04 data set were 29.3\%, 24.5\%, and 19.4\%.

We report the corresponding phone error rates in Table \ref{tab:per}, from which we observe that additional words are indeed resolved as system accuracy increases.  By way of comparison, at the 19.4\% WER system setting, the \emph{oracle} method had a PER of 11.51\%, and the \emph{anti-oracle} had a PER of 27.2\%.  It may also be observed from Table \ref{tab:per} that, for those words which are resolved, the semi-supervised method (\emph{semi-sup}) chooses candidates with smaller edit distance to reference pronunciations from a hand-crafted lexicon.

\subsubsection{Large-Vocabulary Continuous Speech Recognition}

As a final experiment, all four methods described in Section~\ref{sec:ExpProc} for selecting between candidate pronunciations were used to recognize 100 hours of speech that contained all 500 words of interest.  Table \ref{tab:wer} shows a comparison of the results in terms of standard word error rates.  Note that between the two alternative pronunciations, the one with the smaller phoneme edit distance to a reference pronunciation may not necessarily be the one that results in a lower word error rate. Overall, however, a range of about one-half of a percent of WER is observed between the best and worst candidates considered; note from Table \ref{tab:wer} that the supervised selection of pronunciations based on a forced alignment yields a slightly lower error rate in this instance than phoneme edit distance.

Finally, note that the semi-supervised method does as well as the supervised method.  As shown in Table \ref{tab:per}, of the 449 words that were resolved, both the supervised method and the semi-supervised method selected the \emph{same candidate} for 392 of them. Details of the remaining 57 words are presented in Table \ref{tab:57}: Candidate pronunciations are listed in the second and third columns, with the better-performing candidate in bold, and columns 4 and 5 detail the {\it differing} errors due to selecting the candidate pronunciation {\it not in bold} in terms of substitution errors, and insertion/deletion errors.  Many of the words where the methods chose different pronunciations do not impact word error rate---and hence neither is in bold---as the two candidate pronunciations are similar enough that neither results in a lower WER. 

\begin{table}[t]
\center
\begin{tabular}{ccc}
	     {Method} &  {ASR WER\%} & {No. Errors}\\
	     {\it anti-oracle}  &{17.8} & {193,145} \\
  	     {\it sup}  & {17.3} & {187,772}\\
  	     {\it semi-sup}  & {{17.3}} &{{187,424}} \\
          {\it oracle}  & {17.4} & {188,517} \\	
\end{tabular}
\caption{Automatic speech recognition (ASR) word error rates (WER) \label{tab:wer}}
\end{table}

\subsection{Selecting from Amongst $k>2$ Competing Pronunciations}

In practice it may be well necessary to compare more than two pronunciations for a given word.  For example, morphologically rich languages may dictate the consideration of $k>2$ alternative pronunciations for a given orthographic form.  To demonstrate that our techniques remain appropriate in this setting, we adopt here a strategy in which $\binom{k}{2}$ pairwise comparisons are performed for the case $k=3$.  In this approach, every unordered pair of candidate pronunciations is evaluated using the criteria described above for the {\it anti-oracle}, {\it sup}, {\it semi-sup}, and {\it oracle} methods. After all pairwise comparisons have been completed, the candidate chosen the greatest number of times is selected; as noted in Section~\ref{sec:multiway}, a variety of alternative approaches are also possible.

For the results that follow, for each of the 449 words of interest, an additional \emph{third} candidate pronunciation was considered, taken (as the last entry for a given word) from the reference pronunciation lexicon.  Word error rate results for this three-way comparison are shown in Table \ref{tab:wer2}. The {\it anti-oracle} method WER remains the same as in the two-way case (Table~\ref{tab:wer}), as every additional candidate had 0\% PER, and by definition such candidates were not included in the {\it anti-oracle} set. In a similar fashion, the {\it oracle} set contained entirely reference pronunciations. 

Relative to the earlier two-way comparison reported in Table~\ref{tab:wer}, the {\it sup} and {\it semi-sup} sets here contained 288 and 301 new pronunciations, respectively.  The remaining results summarized in Table \ref{tab:wer2} validate the trends observed in the two-way comparison, namely that {\it semi-sup} and {\it sup} perform comparably to each other, as well as to the {\it oracle}. Also, as expected, combining a third pronunciation of high quality resulted in lower error rates for all methods it affected.  

\section{Discussion}
\label{sec:dis}

In showing how censored likelihood ratios may be applied in the context of large-scale speech processing, we have developed in this article a semi-supervised method for selecting pronunciations using \emph{unlabeled data}, and demonstrated that it \emph{performs comparably} to the conventional supervised method. Empirical evidence in support of this conclusion was exhibited across three distinct speech processing tasks that depend upon pronunciation model selection: decision-error trade-off curves for spoken term detection, phone error rates with respect to a hand-crafted reference lexicon, and word error rates in speech recognition.  We have observed these results to be consistent across many words of interest, based on extensive experiments using state-of-the-art systems and well-known data sets.

Note that there are limitations to this method, however, in the context of pronunciation selection.  First, if neither candidate is ever recognized, the ``unconstrained'' recognition step required in the semi-supervised setting can fail to choose a candidate pronunciation for a word. Also, the approach requires having seen textual examples of the word of interest or words like it.  This seems a reasonable requirement, given that a word comes into fashion by being widely noticed. Finally, false alarms in the recognition process may degrade performance---for example, if a word of interest sounds like common word---but our experiments to vary system quality indicated that this problem did not arise for the chosen words of interest in our setting.

In summary, the conventional supervised method for system-level model selection optimizes empirical performance on a labeled development set. Instead, we focused in this article on leveraging \emph{unlabeled data} to choose amongst trained systems through likelihood-ratio-based model selection.  We showed how to generalize the conditional likelihood framework through the use of \emph{automatically generated} labels as a proxy for labels generated by human experts. We then answered the question of how well the resultant censored likelihoods are likely to perform, from both a methodological and an applied perspective.

As a final note, a current research direction of much interest to the speech community attempts to utilize untranscribed utterances for \emph{self-training} of acoustic model parameters~\cite{WessellN, ma}.  While our main interest here was in the general problem of non-nested model selection using unlabeled data, an appealing direction for future work is to take these ideas forward within the acoustic modeling context.

\begin{table}[t]
\center
\begin{tabular}{ccc}
	     {Method} &  {ASR WER\%} & {No. Errors}\\
	     {\it mw-anti-oracle}  &{17.8} & {193,145} \\
  	     {\it mw-sup}  & {17.0} & {184,345}\\
  	     {\it mw-semi-sup}  & {{17.0}} &{{184,297}} \\
          {\it mw-oracle}  & {17.0} & {184,373} \\	
\end{tabular}
\caption{Multi-way (MW) Pronunciation Selection (3 Pronunciations) \label{tab:wer2}}
\end{table}

\section{Acknowledgments}

We gratefully acknowledge the assistance of colleagues at IBM Research and the use of their Attila speech recognition system \cite{IBM}, as well as support and the assistance of colleagues from a sub-team of the 2008 Center for Language and Speech Processing Summer Workshop at Johns Hopkins University, who helped to set up the necessary systems and plan experiments: Abhinav Sethy, Bhuvana Ramabhadran, Erica Cooper, Murat Saraclar, and James K. Baker (co-leader). Also, we would like to acknowledge colleagues in the workshop for providing some of the pronunciation candidates, namely Michael Riley, Martin Jansche, and Arnab Ghoshal.

\bibliographystyle{IEEEtran}%

 \begin{table*}
\center
\begin{tabular}{|c|c|c|c|c|}
            \hline
            Term & \emph{semi-sup} & \emph{sup} & Differing Substitution Errors (No.) & Ins/Del \\
            \hline
            {\it Ahern} & {\bf ey hh er n} & ae er n & ahern $\rightarrow$ upturn (3), apparent (2), hurry (1)  & 6 \\
            {\it Aleve} & {\bf ae l iy v} & ax l eh v & (0) & 1\\
            {\it anybody's} & eh n iy b aa d iy z & eh n iy b ah d iy z & (0) & 0\\
            {\it Asean} & {\bf ax s iy ih n} & ey s iy ih n & asean $\rightarrow$ asham (1) & 2\\
                     &  &  & and $\rightarrow$ asean (1) &\\
            {\it Assuras} & {ax sh uh r ih s} & ax sh uh r ax z & (0) & 0\\
            {\it Avi} & {ax v iy} & ey v iy & (0) & 0\\
            {\it Beatty} & b iy ae t iy & {\bf b ey t iy} & fabiani $\rightarrow$ beatty (1) & 1\\
            {\it Bhuj} & {\bf b uw jh} & b uw zh & bhuj $\rightarrow$ pooch, boost, boots, chip, merge (1) & 5\\
            {\it Canucks} & {\bf k ae n ax k s} & k ae n ah k s & canucks $\rightarrow$ connects (1) & 2\\
                     &  &  & knox $\rightarrow$ canucks (1) &\\
            {\it Cortese} & {\bf k ao r t ey z iy} & k ao r t eh z & cortese $\rightarrow$ he (2), tasty, daisy, taste (1) & 5\\
            {\it Cuellar} & {\bf k w eh l er} & k y uw l er & cuellar $\rightarrow$ korea, out (1) & 2\\
            {\it Dundalk} & {d ah n d ao l k} & d ah n d ao k & (0) & 0\\
            {\it Dura} & {\bf d uw r ax} & d uh r ax & dura $\rightarrow$ dora (1) & 0\\
            {\it Durango}  & {\bf d uh r ae ng g ow} & d uh r ae ng ow & durango $\rightarrow$ tarango (1) & 1\\
            {\it freemen's} & {f r iy m eh n z} & f r iy m ih n z & (0) & 0\\
            {\it Gejdenson} & {g ey hh d ax n s ax n} & g ey hh d ih n s ax n & (0) & 0\\
            {\it Gough}  & {\bf g ao f} & g ao & gough $\rightarrow$ goff (2), damien (1) & 1\\
                     &  &  & schwarzkopf $\rightarrow$ gough (1)$^*$ &\\
            {\it Grosjean}  & {\bf g r ow s jh ih n} & g r ow jh iy n & grosjean $\rightarrow$ are, gross (1), on (1)$^*$ & 1\\
            {\it Hadera}  & {\bf hh ax d eh r ax} & hh ae d eh r ax & hadera $\rightarrow$ era, out (1) & 2\\
            {\it Heupel}  & {\bf hh oy p ax l} & hh y uw p ax l & heupel $\rightarrow$ goals (1) & 1\\
            {\it Ilan}  & {\bf ih l ax n} & ay l ax n & ilan $\rightarrow$ airline (1) & 0\\
            {\it ilo}  & {\bf ay l ow} & ih l ow & ilo $\rightarrow$ iowa, eyal, low (1) & 0\\
            {\it Iverson}  & {\bf ay v er s ax n} & iy v er s ax n & iverson $\rightarrow$ iverson's (14), the (1) & 18\\
            {\it Jonbenet}  & {\bf jh aa n b ax n eh t} & jh aa n b ax n eh & jonbenet $\rightarrow$ they (1) & 1\\
            {\it Jurenovich}  & {\bf jh uw r eh n ax v ih ch} & y uw r eh n ax v ih ch & jurenovich $\rightarrow$ renovate, renovation (3), average (2) & 22\\
                     &  &  & jurenovich $\rightarrow$ events, pitch (2), want (1) &\\
                     &  &  & jurenovich $\rightarrow$ against, batch, each, edge, irrelevant (1) &\\

                     &  &  & jurenovich $\rightarrow$ edge, next, now, sh, tournaments (1) &\\
               {\it Kmart}  & {\bf k ey m aa r t} & k m aa r t & kmart $\rightarrow$ mart (9), answer (2), mark, out (1) & 13\\
                     &  &  & has $\rightarrow$ kmart (1) &\\
            {\it Lampe}  & {l ae m p iy} & l ae m p & (0) & 0\\
            {\it liasson}  & {\bf l y ae s ax n} & ae s ax n & liasson $\rightarrow$ hanson (1) & 1\\
            {\it Likud's}  & {l ih k ah d z} & l ay k uw d z & (0) & 0\\
            {\it Litke}  & {\bf l ih k iy} & l ih t k iy & litke $\rightarrow$ the (1) & 1\\
            {\it Lukashenko}  & {\bf l uw k ae sh eh ng k ow} & l uw k ax sh eh ng k ow & lukashenko $\rightarrow$ i (1) & 1\\
            {\it Marceca}  & {\bf m aa r s ey k ax} & m aa r s eh k ax & marceca $\rightarrow$ because, cut (1) & 1\\
            &  &  & siegel $\rightarrow$ marceca (1)$^*$ &\\
            {\it Matteucci}  & {\bf m ax t ey uw ch iy} & m ae t uw ch iy & matteucci $\rightarrow$ see, to (1), matures (1)$^*$ & 1\\
            {\it Menendez}  & {\bf m eh n eh n d eh z} & m eh n aa n d ey & menendez $\rightarrow$ as (3) & 1\\
                             &  &  & as $\rightarrow$ menendez (3)$^*$ &\\
            {\it Milos}  & {m ay l ow z} & m ih l ow z & (0) & 0\\
            {\it Mustafa}  & {\bf m ah s t ax f ax} & m uw s t aa f ax & mustafa $\rightarrow$ some, sun (1) & 1\\
            {\it Nasrallah}  & {\bf n ae s r aa l ax} & n aa r aa l ax & nasrallah $\rightarrow$ rolla, drama, on (1) & 3\\
            {\it Nhtsa}  & {\bf n ey t s ax} & n t s ax & nhtsa $\rightarrow$ a, nitze (1) & 2\\
            {\it Nkosi}  & {\bf n k ow s iy} & ng k ow z iy & nkosi $\rightarrow$ cozy (1) & 1\\
            {\it Orelon}  &  {ao r l aa n} & ao r ax l aa n & (0) & 0\\
            {\it Ouattara's}  &  {\bf w ax t ae r ax z} & aw ax t ae r ax z & ouattara's $\rightarrow$ tara's (1) & 1\\
            {\it Pawelski}  &  {\bf p ao eh l s k iy} & p ao l s k iy & pawelski $\rightarrow$ belsky, ski (1) & 2\\
            {\it Peltier}  &  {\bf p eh l t iy er} & p eh l t iy ey & peltier $\rightarrow$ tear (2), here, pepsi, years (1) & 5\\
            {\it pre}  &  {\bf p r ax} & p r & pre $\rightarrow$ per (1) & 0\\
            {\it Prodi}  &  {p r ax d iy} & p r aa d iy & (0) & 0\\
            {\it Sadako}  &  {\bf s ax d aa k ow} & s ae d ax k ow & sadako $\rightarrow$ got (1) & 1\\
            {\it Schiavo}  &  {\bf s k y ax v ow} & sh ax v ow & schiavo $\rightarrow$ gavel, ski, elbow, oddball, on, out, will (1) & 1\\
            {\it Schiavone}  &  {\bf s k y ax v ow n} & sh ax v aa n & schiavone $\rightarrow$ bony, bounty (2), a, money, it (1) & 16\\
                             &  &  & schiavone $\rightarrow$ the, voting, about, donate, ioni, owning (1) &\\
            {\it Schlossberg}  &  {sh l ao s b er g} & sh l aa s b er g & (0) & 0\\
            {\it Skurdal}  &  {s k er d ax l} & s k er d aa l & scurbel $\rightarrow$ skurdal (1) & 0\\
                             &  &  & skurdal $\rightarrow$ off (1)$^*$ &\\
            {\it Taliban's}  &  {\bf t ae l ih b ax n z} & t ae l ih b ih n z & metallica $\rightarrow$ taliban's (1) & 1\\
            {\it Thabo}  &  {\bf th aa b ow} & th ax b ow & thabo $\rightarrow$ and, tabor (2) m., problem (1) & 11\\
                             &  &  & thabo $\rightarrow$ hobbled, in, tomlin, trouble, tumbling (1) &\\
            {\it tornados}  &  {\bf t er n ey d ow z} & t ao r n ey d ow s & (0) & 0\\
            {\it Yasir}  &  {y ax s iy r} &{\bf y aa s iy r} & yasir $\rightarrow$ oster (1) & 1\\
            {\it Yugoslavs}  &  {\bf y uw g ow s l aa v z} & y uw g ow s l aa v s & (0) & 0\\
            {\it Zhirinovsky}  &  {\bf zh ih r ih n ao v s k iy} & iy r ih n ao v s k iy & zhirinovsky $\rightarrow$ ski, skin, speak (1) & 3\\
            {\it Zorich}  &  {\bf z ax r ih ch} & z ow r ih k & zorich $\rightarrow$ storage, h., is (2) & 6\\
\hline
\end{tabular}
\caption{Words where the methods differ in selection. Differing errors listed caused by the non-bold pronunciation marked with an $*$. \label{tab:57}}
\end{table*}

\end{document}